\documentclass[conference]{IEEEtran}
\IEEEoverridecommandlockouts

\usepackage{cite}
\usepackage{amsmath,amssymb,amsfonts}
\usepackage{algorithmic}
\usepackage{graphicx}
\usepackage{textcomp}
\usepackage{xcolor}
\usepackage{booktabs}
\usepackage{multirow}
\usepackage{url}
\usepackage[table]{xcolor}
\usepackage{array}
\usepackage{subcaption}
\usepackage{stfloats}
\usepackage{hyperref}

\def\BibTeX{{\rm B\kern-.05em{\sc i\kern-.025em b}\kern-.08em
    T\kern-.1667em\lower.7ex\hbox{E}\kern-.125emX}}

\begin{document}

\title{SPEARBench: A Benchmark for Naturalness Evaluation in Streaming Speech-to-Speech Language Models
}

\author{Anonymized submission to IEEE SLT 2026}
\author{\IEEEauthorblockN{Thomas Thebaud\IEEEauthorrefmark{1}, Yuzhe Wang\IEEEauthorrefmark{1}, Hao Zhang\IEEEauthorrefmark{1}, Sathvik Manikantan Napa Ugandhar\IEEEauthorrefmark{1}, Ashish Hallur\IEEEauthorrefmark{1},\\ 
Georgi Tinchev\IEEEauthorrefmark{2}, Venkatesh Ravichandran\IEEEauthorrefmark{2}, Laureano Moro-Velazquez\IEEEauthorrefmark{1}}
\IEEEauthorblockA{\IEEEauthorrefmark{1}\textit{Department of Electrical and Computer Engineering, Johns Hopkins University}, Baltimore, MD, USA\\
Email: tthebau1@jhu.edu}
\IEEEauthorblockA{\IEEEauthorrefmark{2}\textit{Amazon Inc.}, Seattle, WA, USA}}

\maketitle

\begin{abstract}
Streaming speech-to-speech language models aim to answer spoken queries directly with synthetic speech. However, standard speech and text benchmarks do not capture whether these systems behave naturally in conversations, where timing, turn-taking, prosody, interpersonal stance, language and dialect consistency, and relationship-aware appropriateness jointly shape perceived quality. We introduce SPEARBench, a benchmark for evaluating naturalness in speech-to-speech language models from question-answer interactions. SPEARBench constructs controlled dialogue prompts from the Seamless Interaction corpus, runs inference across multiple models, and evaluates generated answers using a multidimensional protocol that covers response latency, interruptions, speech quality, ASR robustness, language and dialect consistency, emotional naturalness, interpersonal stance, and explainable distributional baselines. The benchmark includes original human answers as a reference condition and reports results for several contemporary models. 
Results show that current models can achieve high signal-level quality and low ASR error while still differing from human conversational behavior in latency, overlap, dialect preservation, emotional adaptation, and interpersonal stance dynamics.
\end{abstract}


\section{Introduction}
Large language models have recently become central to natural language processing, human-computer interaction, and artificial intelligence more broadly~\cite{annepaka2025large}.
Their success has motivated spoken language models that extend text-based interaction to speech. In this paper, we focus on speech-to-speech (S2S) language models, defined as systems that take speech as input and produce speech as output, enabling interaction without an explicit text interface~\cite{arora2025landscape}.
Some recent systems operate in streaming mode, producing answers with low enough latency for real-time interaction, and some are full-duplex systems, meaning that they can process incoming speech while simultaneously producing outgoing speech~\cite{wang2024full}.
Together, these capabilities make natural spoken conversation with a model increasingly realistic.
However, many current systems still rely on encoding, processing, and decoding pipelines that can remove paralinguistic cues or limit synthesized-speech flexibility, reducing conversational naturalness.
Human conversational naturalness is shaped by coupled phenomena including response timing, pauses, overlap, interruptions, prosody, affect, stance, language choice, dialect consistency, context, and speaker relationship~\cite{dunbar1997human,pomerantz20046}.
A S2S model that produces intelligible, high-quality audio may still feel unnatural if it responds too quickly, interrupts the user, shifts dialect unexpectedly, flattens emotional dynamics, or answers with an inappropriate voice. 

Existing spoken dialogue and speech generation evaluations provide valuable tools for some aspects of naturalness evaluation~\cite{ray2026tau, zhang2025wildspeech}.
However, most existing benchmarks either require human annotators, which makes evaluation costly, or focus on one aspect of the conversation, such as latency~\cite{ray2026tau}, or interruptions~\cite{lin2026full}.
We present SPEARBench, a benchmark for naturalness evaluation in conversational S2S systems, and an online platform for sharing model performance across a unified set of metrics. 
SPEARBench is built on two-speaker question-answer evaluation clips from the Seamless Interaction dataset~\cite{agrawal2025seamless}.
The benchmark extracts contexts in which one speaker asks a question and the next speaker answers, then compares the original human answer with generated answers from streaming S2S language models. 
The pipeline computes answer-level acoustic, textual, temporal, linguistic, emotional, stance, turn-taking naturalness, and explainable distributional metrics using a set of state-of-the-art (SOTA) models and aggregates them into compact benchmark reports.

This work contributions can be summarized as:
\begin{itemize}
    \item Defining a reproducible benchmark protocol for evaluating naturalness in S2S language models using controlled conversational prompts. The dataset extracted from Seamless Interaction~\cite{agrawal2025seamless} is available for download\footnote{Dataset available in the "Add your model" page of the website.}, as well as the code to apply this protocol to new evaluation datasets.
    \item Integrating a set of validated metrics and models for emotional naturalness~\cite{TRACE}, interpersonal stances~\cite{STANCE}, turn-taking naturalness~\cite{TURNS}, speech emotion recognition~\cite{feng2025vox}, language and dialect recognition~\cite{feng2026voxlect}, and Conversational baselines~\cite{DISTRIB} into a single publicly available evaluation pipeline\footnote{GitHub: \url{https://anonymous.4open.science/r/SPEAR-benchmark-code-anon-F4F1}}.
    \item Evaluating multiple S2S systems, highlighting discrepancies between high audio quality and human-like conversational behavior. Summarized and detailed results can be read in a publicly available website, which will gain new models over time.\footnote{Website: \url{https://anonymous.4open.science/w/SPEAR-benchmark-website-anon-82FE}}
\end{itemize}

\section{Related Work}
\label{sec:related_work}

\subsection{Streaming Speech-to-Speech Language Models}
\label{subsec:rw_S2S_models}
Text-based large language models have become central tools for language understanding, reasoning, and generation~\cite{openai2024gpt4technicalreport, xu2025qwen3, team2023gemini}.
Speech-aware language models extend these capabilities to spoken inputs by using speech tokenizers~\cite{zhang2023speechtokenizer}, ASR models~\cite{gong2023whisper, shi2026qwen3}, or speech foundation models~\cite{hsu2021hubert, wav2vec2_2020, chen2022wavlm} to connect continuous audio with language-model representations~\cite{hu2024wavllm, chu2023qwen}. 
S2S models go further by taking speech as input and producing speech as output, as in \textit{GPT-audio-1.5}~\cite{GPT-audio-1.5}, preserving information that may be lost in text, such as emotion, style, timing, and prosody. 
Streaming models such as \textit{Qwen3-Omni}~\cite{xu2025qwen3}, \textit{Qwen2.5-Omni}~\cite{Qwen2.5-Omni}, \textit{GPT-realtime-2}~\cite{GPT-realtime-2}, \textit{Gemini flash live} 2.5~\cite{comanici2025gemini} and 3.1~\cite{gemini-3.1-flash-live-preview}, and \textit{Mini-Omni}~\cite{xie2024mini} process audio incrementally and can enable streaming full-duplex interactions. 

\subsection{Conversational Naturalness Evaluation}
\label{subsec:rw_naturalness_metrics}

Human speakers adapt timing, pitch, rhythm, emotion, and stance to the conversational context, while prosody carries information about emphasis, emotion, engagement, and turn structure~\cite{couper1996prosody, walker2012phonetics}. 
They also show entrainment across acoustic, prosodic, lexical, and temporal features, which has been studied as a marker of coordination and conversational quality~\cite{wynn2022classifying}. 
Natural conversation therefore requires not only intelligible speech, but also appropriate timing and turn-taking~\cite{brusco2020cross}: late responses may feel slow, while early responses may interrupt the user, especially in full-duplex systems that must decide when to answer while speech is still arriving~\cite{smith2015real, arora2025talking}. 

Standard metrics such as WER and UTMOS~\cite{saeki2022utmos} remain useful, but do not fully capture these interactional dimensions: a model can have low WER and high speech quality while answering at the wrong time, using a flat voice, missing the emotion of the question, or changing the interpersonal stance. 
Recent work has demonstrated that evaluator models can be used to reliably estimate conversation-level properties, including turn-taking naturalness with Voice Activity Prediction models~\cite{TURNS}, emotional naturalness and entrainment~\cite{TRACE}, interpersonal stance~\cite{STANCE}, and dialect profiles across languages~\cite{feng2026voxlect}. 
SPEARBench leverages these models in Section~\ref{subsec:eval}.

\subsection{Existing Benchmarks for Speech-to-Speech Language Models Evaluation}
\label{subsec:rw_existing_benchmarks}

Many benchmarks evaluate speech processing models and LLMs. 
Some focus on speech understanding tasks such as transcription, spoken question answering, speech translation, or audio reasoning, including SUPERB-style benchmarks for speech foundation models~\cite{yang2021superb}. 
Others compare LLM abilities in text-only~\cite{lin2025wildbench,bai2024mt} or speech-based settings~\cite{yang2024air,zhang2025wildspeech,tan2026globeaudio}. 
These benchmarks measure important capabilities, but they often focus on isolated utterances or task correctness rather than conversational behavior, leading to known gaps and inconsistencies across evaluations~\cite{laskar2024systematic}. 
Recent work has begun evaluating full-duplex spoken interaction, including emergency scenarios~\cite{ge2025flexi}, latency~\cite{ray2026tau}, interruptions~\cite{lin2026full}, and turn efficiency~\cite{ge2025flexi}. Some commercial initiatives, such as Artificial Analysis \cite{artificialanalysis}, provide scores for AI agents that include automatic evaluation of pause handling, turn-taking, interruption handling, and backchannel handling, but they do not provide an open-source benchmark with peer-reviewed reports or articles. Other initiatives, such as Speech Arena \cite{speecharena_leaderboard} or Voice Arena \cite{voicearena_tts_methodology} are more focused on text-to-speech systems rather than AI agents and mainly use human evaluations to create their leaderboards.
Consequently, most benchmarks still focus on a single dimension, are not open, or require human intervention. 
SPEARBench complements them by providing an open-source automated benchmark for conversational naturalness on a wide set of aspects.



\begin{figure}[t]
    \centering
    \includegraphics[width=\linewidth]{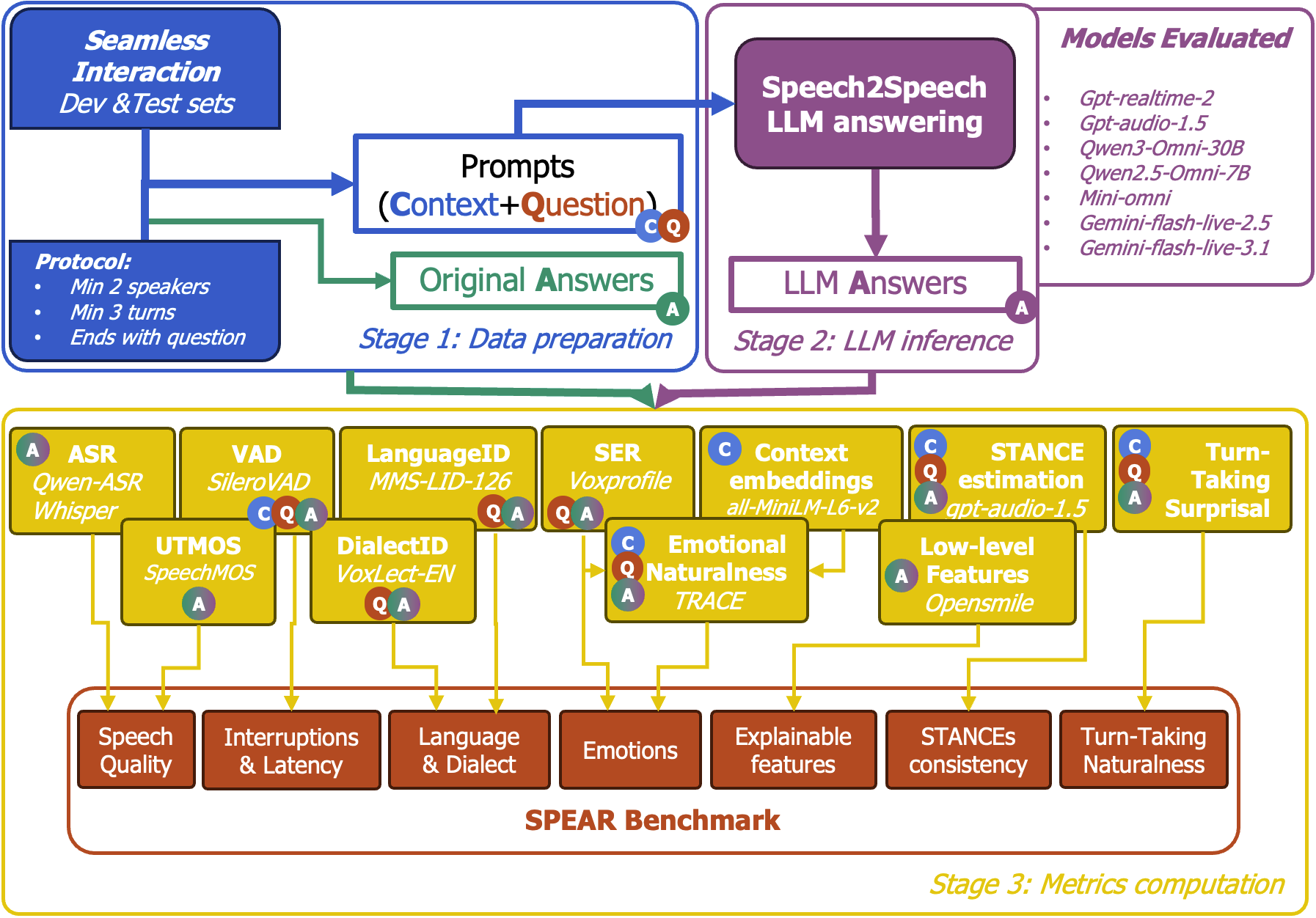}
    \caption{SPEARBench pipeline.}
    \label{fig:method}
    \vspace{-5mm}
\end{figure}

\section{Methods}
This section details the proposed benchmark implementation, which consists of three steps: data preparation, LLM inference for each S2S model evaluated, and then evaluation.
Figure~\ref{fig:method} summarizes the pipeline of SPEARBench. 

\subsection{Data Preparation}

\subsubsection{Dataset}
In this work, we use the dev and test sets of the Seamless Interaction dataset~\cite{agrawal2025seamless}, a collection of dyadic spoken conversations of up to 10 minutes each, designed to capture realistic human-human interaction. 
Both the \textit{improvised} and \textit{naturalistic} subsets of the corpus are used. 
The \textit{naturalistic} subset contains prompted conversations carried out by non-actor participants and is intended to capture realistic everyday interaction patterns, while the \textit{improvised} subset contains scenario-based interactions, and provides more controlled conversational situations designed to elicit particular roles or emotions, including a manual annotation of the interpersonal stances of the speakers, as well as their relationships.
All conversations have been transcribed and diarized, and audio for each speaker is recorded on 2 different channels.
Speakers include a range of ethnicities, accents, ages, and levels of education, as well as a gender balance close to parity.
Table \ref{tab:dataset} shows the original size of the dev and test sets before and after dialogue selection.

\begin{table}[ht]
    \centering
    \caption{Number and hours of audios for the original conversations, and the selected dialogues.}
    \begin{tabular}{l llll l}
    \toprule
         & \multicolumn{2}{c}{Improvised} & \multicolumn{2}{c}{Naturalistic} & \multirow{2}{*}{Total}\\
        \cmidrule(r){2-3} \cmidrule(r){4-5}
        & Dev & Test & Dev & Test & \\
    \midrule
    Original conversations  & 342 & 232 & 691 & 742 & 2007 \\ 
    Original hours      & 21.81h & 15.84h & 35.33h & 40.82h & 113.80h \\ 
    \midrule
    Selected dialogues  & 1165 & 994 & 1529 & 1731 & 5419 \\ 
    Contexts+Questions  & 4.78h & 3.76h & 6.76h & 7.47h & 22.77h \\ 
    Answers             & 2.67h & 2.29h & 5.23h & 4.36h & 14.56h \\ 
    Total selected hours & 7.45h & 6.05h & 12.00h & 11.83h & 37.33h \\ 
    \bottomrule

    \end{tabular}
    \vspace{-5mm}
    \label{tab:dataset}
\end{table}

\subsubsection{Dialogue selection}
A conversation from $N$ speakers $\{S_n|n\in[\![1,N]\!]\}$ can be represented as a set of $K$ turns,
We define one turn as a succession of utterances from a given speaker without interruption of backchannel over 0.5 seconds, so by definition, two consecutive turns are from different speakers.
To extract one \textit{valid dialogue} from a recorded conversation, we search for any set of $K\geq 3$ consecutive turns from $N=2$ speakers, whose second-to-last turn contains a question (transcription ends with a '\texttt{?}').

This creates three parts in any given valid dialogue: 
\begin{enumerate}
    \item \textbf{Context}: turns $[1,K-2]$, show the previous exchanges in the conversation, from speakers $\{S_1,S_2\}$.
    \item \textbf{Question}: turn $K-1$, shows an inquiry made by speaker $S_1$, which we know had an existing answer by a real speaker. The constraint imposed on the question is the simplest condition that ensures an answer. 
    \item \textbf{Answer}: turn $K$, shows the human answer given by a speaker $S_2$, for later comparison to the evaluated models.
\end{enumerate}

This selection protocol identified at least one \textit{valid dialogue} in 70.91\% of the conversations, with an average of 4.6 valid dialogues extracted per conversation.
Table \ref{tab:dataset} shows the size of the dev and test sets after dialogue selection.

\subsection{LLM inference}
For each selected interaction, the evaluated model receives the context and question audio as input and is asked to produce a spoken answer. 
The benchmark stores two outputs for each inference: the generated answer waveform and the estimated start time of the answer.
The start time is computed as the difference between the absolute time at which the model begins producing speech and the end of the context+question input. 
A positive value, therefore, indicates that the model waited until the end of the question before answering, while a negative value indicates that the model began answering before the input audio ended. 
Negative start times are treated as interruptions, since the model starts speaking before the human speaker has finished.
We evaluate seven open-weights and proprietary S2S models with a range of model sizes and inference regimes, shown in Table \ref{tab:models}
\begin{table}[bh]
    \centering
    \vspace{-5mm}
    \caption{Models evaluated, with their inference mode, number of parameters when known, and }
    \resizebox{\linewidth}{!}{%
    \begin{tabular}{llc}
    \toprule
    Model & Inference & Open\\
    (\# parameters when known) & mode & weights \\
    \midrule
    \texttt{GPT-audio-1.5}~\cite{GPT-audio-1.5} & Non-streaming \\
    \texttt{GPT-realtime-2}~\cite{GPT-realtime-2} & Full-duplex \\
    \texttt{Qwen3-Omni-30B-Instruct}~\cite{xu2025qwen3} (30B) & Full-duplex & $\checkmark$\\
    \texttt{Qwen2.5-Omni-7B}~\cite{Qwen2.5-Omni} (7B)& Half-duplex & $\checkmark$\\
    \texttt{Gemini-2.5-flash-native-audio}~\cite{comanici2025gemini} & Full-duplex \\
    \texttt{Gemini-3.1-flash-live}~\cite{gemini-3.1-flash-live-preview} & Full-duplex \\
    \texttt{Mini-omni}~\cite{xie2024mini} (0.5B) & Half-duplex & $\checkmark$\\
    \bottomrule
    \end{tabular}}
    \label{tab:models}
\end{table}


\subsection{Evaluation}
\label{subsec:eval}

\subsubsection{Intelligibility and Speech Quality}
We first include a small set of traditional metrics to facilitate comparison with existing work: \textit{intelligibility} and \textit{speech quality}.
\textbf{Intelligibility} is measured through the relative success of Automatic Speech Recognition (ASR) systems.
The transcripts of the answers are computed using a set of ASR systems (\texttt{whisper-large-v3}~\cite{radford2023robust}, and \texttt{qwen3-ASR-0.6B}~\cite{shi2026qwen3}).
For each answer, the WER and Character Error Rate (CER) are measured for each system's outputs to assess the models' intelligibility relative to human intelligibility in real-life scenarios, while reducing bias from a single evaluator.
All the S2S models evaluated in this study provide a text version of their outputs upon inference, which are used as a reference to calculate WER.
\textbf{Speech Quality} is measured through the Mean Opinion Score (MOS), automatically evaluated by the UTMOS model~\cite{saeki2022utmos}.

\subsubsection{Interruptions and Latency}
\textbf{Latency} is defined as the time between the end of the question's speech and the beginning of the voiced answer, measured using the Silero VAD~\cite{SileroVAD} Voice Activity Detection model.
Cases of interruption are not counted toward the latency, as they would theoretically have a negative latency.
In this context, we define \textbf{time of interruption} as the total duration during which the model's speech overlaps with the original speaker's speech.
If the \textit{time of interruption} is not zero, one interruption is counted, being the total \textbf{number of interruptions} reported as well.

\subsubsection{Turn-Taking Naturalness}
\label{subsec:turns}
We estimate turn-taking naturalness using predictive turn-taking models trained on natural dyadic conversations. 
Following the formulation of turn-taking naturalness as the predictive typicality of future voice activity in a two-speaker dialogue, we score each reconstructed interaction with the best evaluator model from~\cite{TURNS}.
This evaluator is based on the DualTurn-based model, which predicts future voice activity from the two-channel dialogue signal and assigns higher surprisal to less plausible local turn-taking patterns. For each model answer, we compute the turn-taking naturalness score for the answer given the context and the question, with both evaluator families, and average their outputs to reduce evaluator-specific bias.

\subsubsection{Language and Dialect Consistency}
\label{subsec:dialect}
\textbf{Language} is predicted using the MMS language identification system~\cite{pratap2023mms}, trained on 4,017 languages, providing up to 97.3\% classification accuracy over 102 languages when evaluated on the FLEURS benchmark~\cite{conneau2023fleurs}.
As Seamless Interaction is an English-only dataset, only the percentages of English and non-English answers are reported, as an answer in a different language, if not explicitly prompted, would be considered a failure.
\textbf{Dialect profile} of the question and the answer is computed using \texttt{Voxlect-English}~\cite{feng2026voxlect}, which shows a 83.0\% accuracy over 16 English dialect categories: East Asia, English, Germanic, Irish, North America, Northern Irish, Oceania, Welsh, Romance, Scottish, Semitic, Slavic, South African, Southeast Asia, South Asia, and Others.
Irish, Northern Irish, Scottish, English, and Welsh are merged into the '\texttt{British Isles}' category; '\texttt{Others}' is ignored.
The resulting vector of 11 logits is used as a representation of the dialect profile.
We use those dialect profiles to measure the \textbf{dialectal entrainment} as the \textit{least squares regression coefficient} between the dialect of the question and the answer, and to measure the \textbf{dialectal variance} as the \textit{total variance} (trace of the covariance matrix) of the dialect profiles. In this sense, an agent changing dialects during a conversation (measured through \textbf{dialectal variance}) could be considered unnatural. In contrast, the \textbf{dialectal entrainment} metric provides information about the dialectal alignment between a speaker and the S2S model providing a response, which can be useful in many scenarios, but it is not always linked to naturalness, as a conversation between two speakers with different dialects can sound highly natural.

\subsubsection{Emotional Naturalness and Entrainement}
\label{subsec:emotion}
The emotional naturalness stage leverages the \textbf{TRACE} model~\cite{TRACE}, which has shown a 97.01\% accuracy to classify the naturalness of conversation based on their emotions on real and simulated datasets. 
For each pair of question and answer, embeddings are extracted using the Voxprofile Speech Emotion Recognition model~\cite{feng2025vox}, as well as the Arousal, Valence and Dominance (AVD) of each utterance. 
Conversation context embeddings are extracted from the conversation prompts given to the participants, using the SBERT model \texttt{all-MiniLM-L6-v2}~\cite{reimers-2019-sentence-bert}.
Relationship embeddings are one-hot vectors computed from the relationships reported in the dataset.
Emotion, context, and relationship embeddings are then input to the pre-trained emotional naturalness predictor~\cite{TRACE}, which outputs an \textbf{emotional naturalness score}.
In this study, values should be interpreted relative to the original-human-answer condition and across evaluated systems rather than as an absolute perceptual score.
Correlation coefficients between the AVD values of the question and the answer are also reported, allowing a comparison of the emotional entrainment between the original dialogues and the LLM's answers.

\subsubsection{Interpersonal Stances}
\label{subsec:stances}

For the improvised subset, SPEARBench evaluates whether each generated answer preserves the interpersonal behavior of the original human answer in the same question context.
Following the StanceBench methodology, an audio-language model judge rates the answer in the question context along 10 stance dimensions: politeness, empathy, warmth, calmness, deference, sociability, organization, focus, honesty and disinhibition. 
The chosen judge (\texttt{GPT-audio}~\cite{GPT-audio-1.5}) has shown among the highest AUROC values across evaluated models for all stances~\cite{STANCE} (average of 0.83 AUROC).
We compare the generated answer with the corresponding question and report the percentage of cases in which the generated answer has  the \textbf{same stance as the question}, is \textbf{more negative}, or \textbf{more positive}~\cite{STANCE}. 
Here, positive and negative refer to the positive and negative poles of each dimension defined in StanceBench, such as warm versus cold, polite versus rude or focused versus withdrawn.

\subsubsection{Temporal and Spectral Features}
SPEARBench extracts spectral and temporal features following a previous study on explainable distributional baselines on natural conversations~\cite{DISTRIB}.
For spectral aspects: Pitch values are extracted and normalized per speaker to measure the amplitude and variations per decile for each S2S model and for the human answers. 
For temporal aspects: The durations of the answers, the voiced time, and the ratio of voiced time are also measured and compared against the human reference. 
The benchmark reports the average duration metrics on the test subsets, as well as the average standard deviation of the normalized pitch values.

\vspace{-3mm}
\section{Results}


\subsection{Intelligibility and Speech Quality Results}
\label{subsec:intelligibitily_results}

The section '\textbf{Speech quality}' of Table~\ref{tab:results} shows the WER, CER, and UTMOS for all evaluated models.
We observe that the generated speech is often more intelligible than human speech, with a slight degradation in older models like \texttt{Mini-omni} and an expected improvement in non-streaming models like \texttt{GPT-audio-1.5}.


\vspace{-2mm}
\subsection{Interruptions and Latency Results}
\label{subsec:interruption_results}

We measure response latency as the delay between the end of the question and the start of the model answer. 
The section '\textbf{Interruptions}' of Table~\ref{tab:results} reports the average latency, the percentage of interrupted samples, and the average interruption duration. 
These metrics allow us to distinguish models that answer quickly but interrupt too often from models that wait for the speaker to finish but produce delayed responses.
The latencies shown by all the models (in the 959-2724ms range) are slightly longer than the average human answer time (742ms in this dataset), which may either sound unnatural to humans or be perceived as a lack of engagement or focus, which could be correlated with the stance of the models being perceived as lower focus than the humans, as shown in Section \ref{subsec:stances_results}.
Most full-duplex models exhibit higher latency than half-duplex models.
Most models show no interruption at all, with the exception of \texttt{GPT-realtime-2} (23.6\% of dialogues interrupted, using a \textit{semantic-vad}), and \textit{mini-omni} (58.1\% of dialogues interrupted). 
We observe that the average interruption time of \texttt{GPT-realtime-2} remains low at 431ms, indicating a limited impact on the conversation, while the one from \textit{mini-omni} way higher, reaching 1243ms.

\begin{figure}[t]
\vspace{-5mm}
    \centering
    \includegraphics[width=\linewidth]{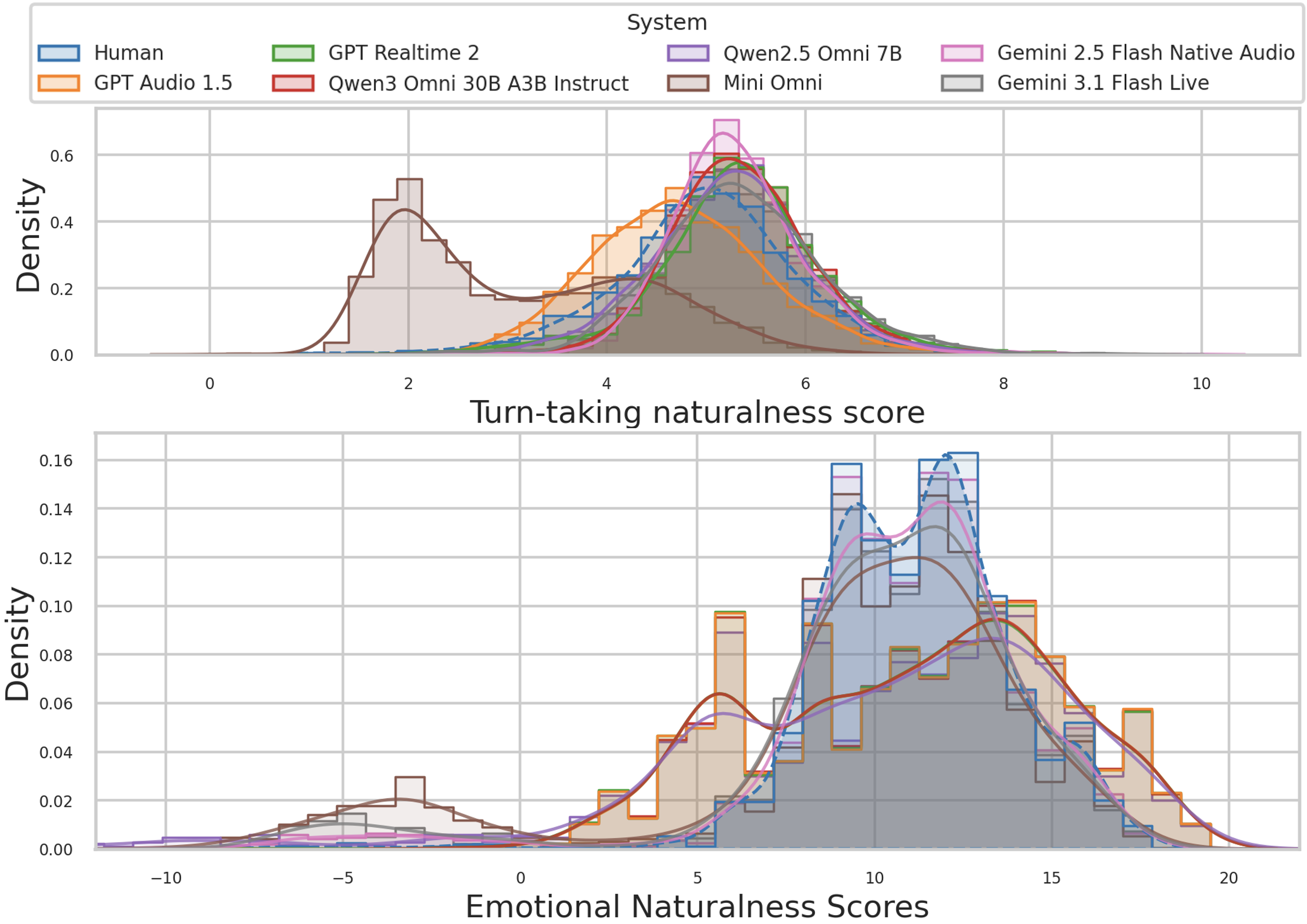}
    \caption{Histograms showing the distribution of the turn-taking naturalness scores (top) and the emotional naturalness scores (bottom).}
    \label{fig:S4}
    \vspace{-7mm}
\end{figure}

\subsection{Turn-Taking Naturalness Results}
The turn-taking naturalness average scores are reported in the \textbf{Turn-Taking} section of Table~\ref{tab:results}, while their distribution is shown in Figure~\ref{fig:S4} (top). 
Most recent full-duplex models obtain scores close to, or slightly higher than, the human reference, suggesting that their global voice-activity patterns are generally compatible with natural dyadic turn-taking. 
In contrast, \texttt{Mini-Omni} receives a much lower score, consistent with its higher interruption rate and less stable conversational timing. 
These results indicate that turn-taking naturalness captures information complementary to simple latency and interruption metrics: a model can answer quickly and rarely overlap with the speaker while still producing timing patterns that differ from those of natural conversation.

\begin{figure}[b]
    \vspace{-5mm}
    \centering
    \includegraphics[width=\linewidth]{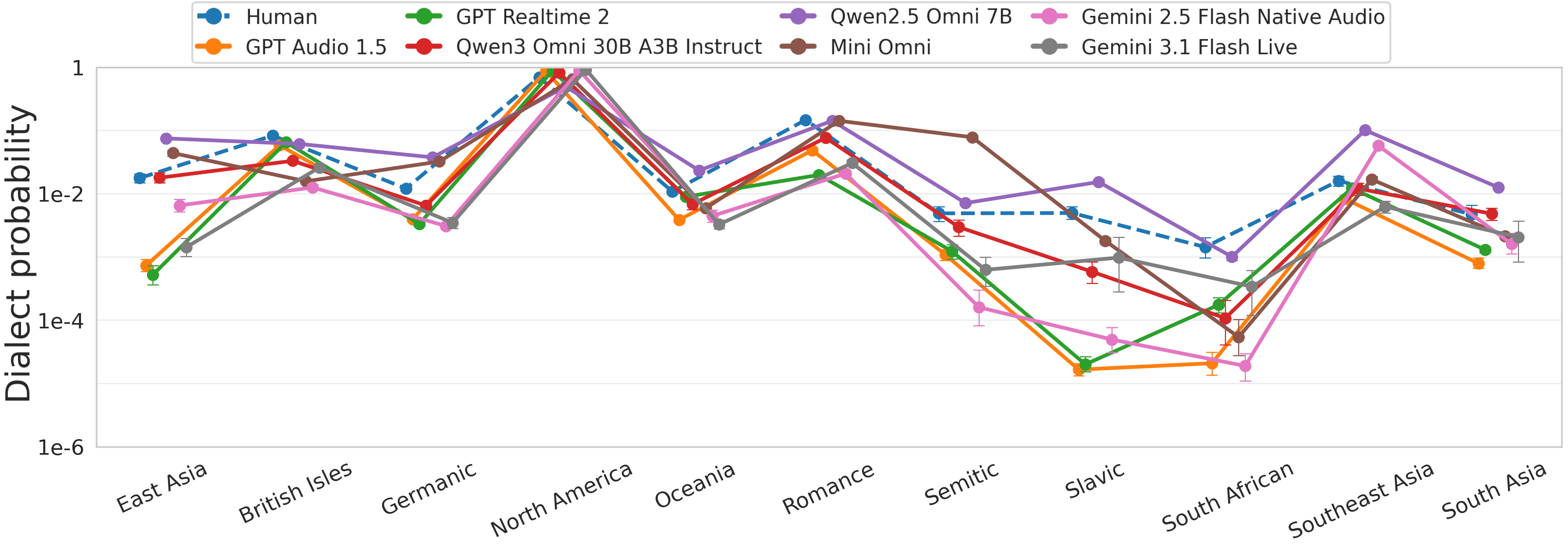}
    \caption{Dialect profiles. This pointplot shows the average dialect profile per model.}
    \label{fig:S3}
    \vspace{-7mm}
\end{figure}

\subsection{Language and Dialect Consistency Results}
\label{subsec:language_results}

The section '\textbf{Language and Dialect}' of Table \ref{tab:results} shows the results related to this section.
As expected, the vast majority of answers to English questions are detected as English (in the 98.1-100\% range).
The other languages detected correspond to a long tail of languages, which, given the 97.3\% accuracy of the language ID model used, may be attributed to either the evaluator model's error or to LLM answers that mismatch the target language.
Figure~\ref{fig:S3} compares the average dialect profiles of all the systems in a pointplot using the log of the logits. 
First, we can see that most models have a similar profile (a strong North American dialect prior, and clear contributions from the British Isles and Romance groups of dialects), which may be linked to internal biases in the dialect id model. 
Only \texttt{Qwen2.5} and \texttt{Mini-omni} seem to present a very different profile, much stronger in the Asian dialects and, respectively, the Slavic and Semitic dialects. 
To measure the dispersion of the model's dialect profiles, we use the trace of the covariance matrix of the profile vectors, reported in \ref{tab:results} as '\textit{dialectal variance}'.

Additionally, we measured the dialectal entrainment between the questions and the answers, and as shown in Table \ref{tab:results}, there is little to no correlation between the questions and the answers dialects, showing that none of the evaluated models currently follow the questions dialect.

\subsection{Emotional Naturalness and Entrainement Results}
\label{subsec:emotion_results}

Conversational naturalness also depends on whether the answer is emotionally appropriate for the context. 
We evaluate this with TRACE, which combines speech emotion representations from the question and answer with contextual and relationship information. 
The average emotional naturalness scores are reported in the \textbf{Emotions} section of Table~\ref{tab:results}, and their distributions are shown in Figure~\ref{fig:S4} (bottom). 
Human answers obtain the highest scores, followed closely by most models, while \texttt{Qwen2.5} and especially \texttt{Mini-Omni} show a thicker low-score tail, suggesting a weaker emotional fit for some responses.
Overall, all models show an emotional naturalness score significantly lower than humans (p-value $<10^{-13}$ for all models using a paired Friedman test).

\begin{figure}[t]
    \vspace{-5mm}
    \centering
    \includegraphics[width=\linewidth]{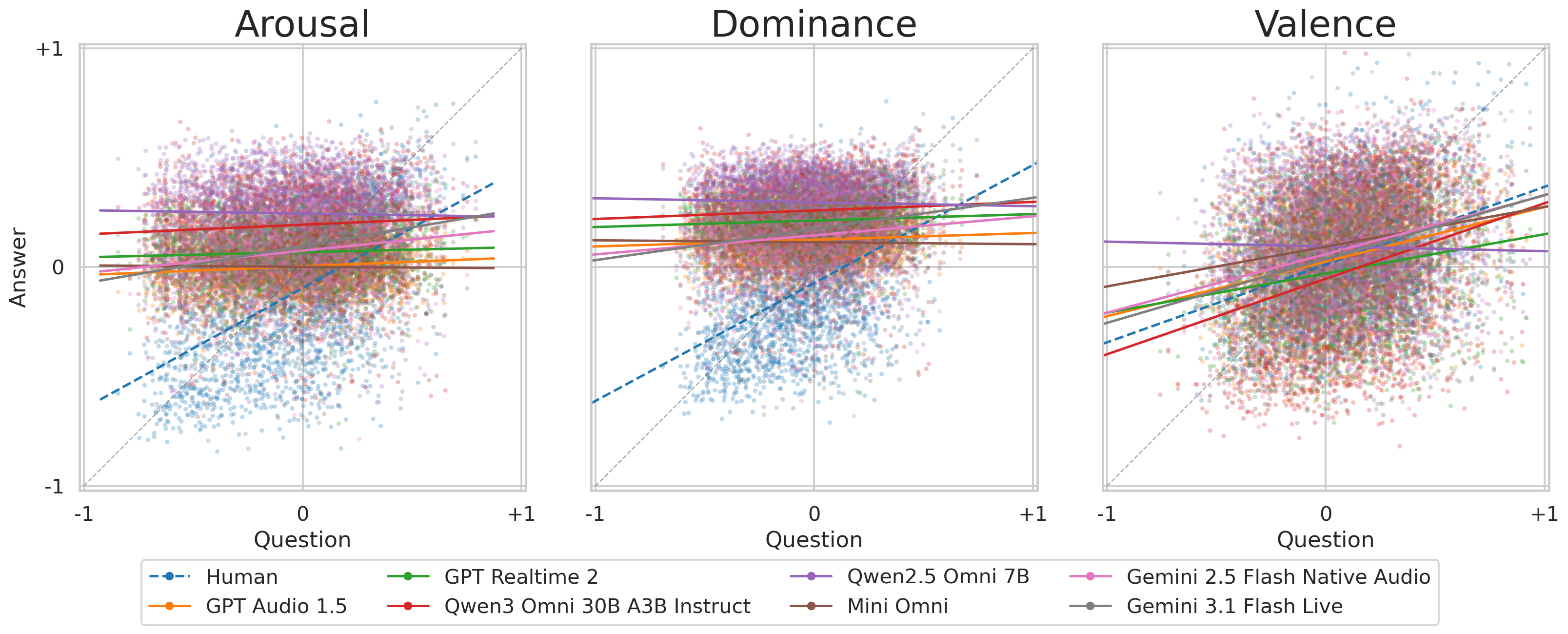}
    \caption{Scatter plots of the Arousal/Valence/Dominance predicted value for the answer compared to the question, and plots of the correlation lines.}
    \label{fig:S5}
    \vspace{-5mm}
\end{figure}

We also analyze emotional entrainment using arousal, valence, and dominance estimates for the question and answer turns. 
Figure~\ref{fig:S5} shows the corresponding question-answer scatter plots, and Table~\ref{tab:results} reports their correlations. 
Human answers show clear correlations across all three dimensions, while model answers mostly preserve valence and show little correlation for arousal or dominance. This suggests that current models can often match the general emotional polarity of the question, but do not fully adapt their emotional intensity or dominance to the conversational context.

\begin{figure}[b]
    \centering
    \includegraphics[width=\linewidth]{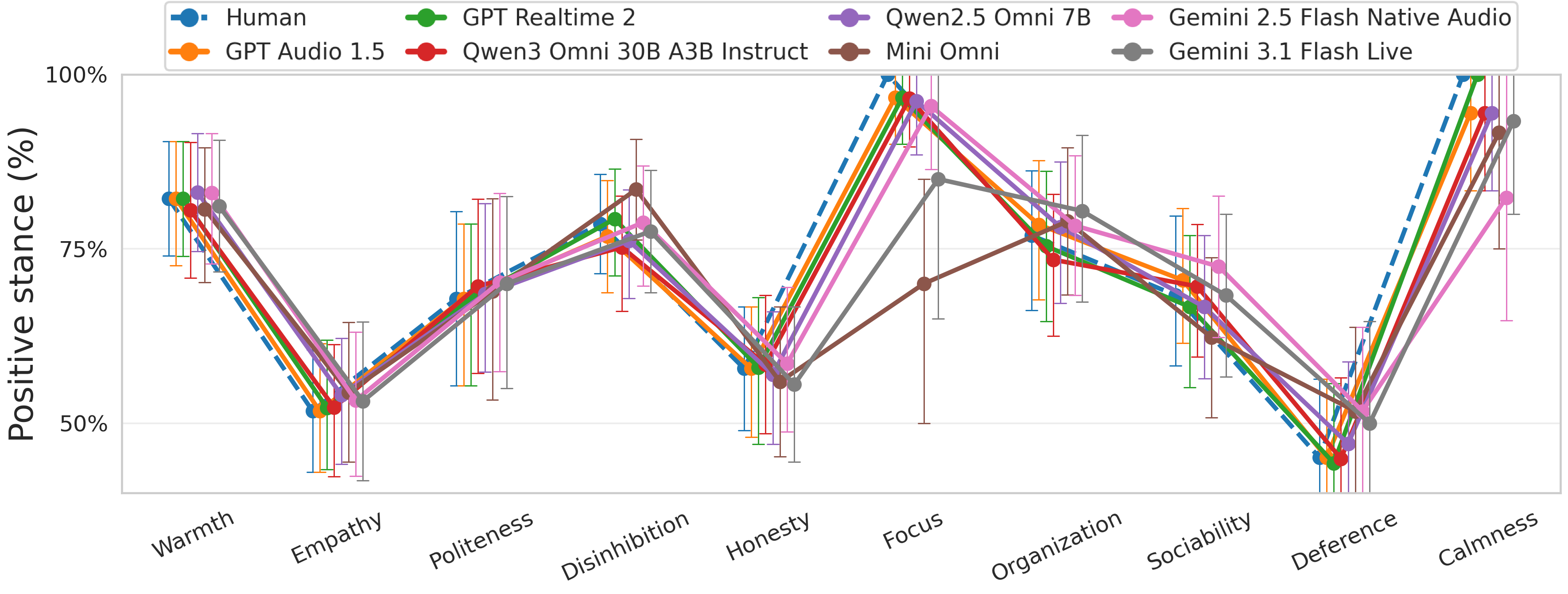}
    \caption{Interpersonal stance results. The pointplot reports the percentage of positive stance answers per model.}
    \label{fig:S6}
    \vspace{-5mm}
\end{figure}

\subsection{Interpersonal Stances Results}
\label{subsec:stances_results}



We evaluate interpersonal stance on the \textit{improvised} subset, assessing whether the evaluated models' responses maintain the same interpersonal behavior as the original human responses.
Figure~\ref{fig:S6} shows the average stance of each model on a point plot.
Most models closely follow the human answers' stances, except that \texttt{Mini-omni}, which shows much lower \textit{Focus}.
We summarize stance preservation as the percentage of responses with the same, more positive, or more negative stance than the question, aggregated across all stance dimensions.
Those results, reported in the section '\textbf{Stances}' of Table~\ref{tab:results}, show that all models closely preserve the human answer's interpersonal stance, with the same-stance percentage above 95\% across all models.

\subsection{Temporal and Spectral Features Results}
\label{subsec:explainable_results}

\begin{figure}[h]
    \centering
    \includegraphics[width=\linewidth]{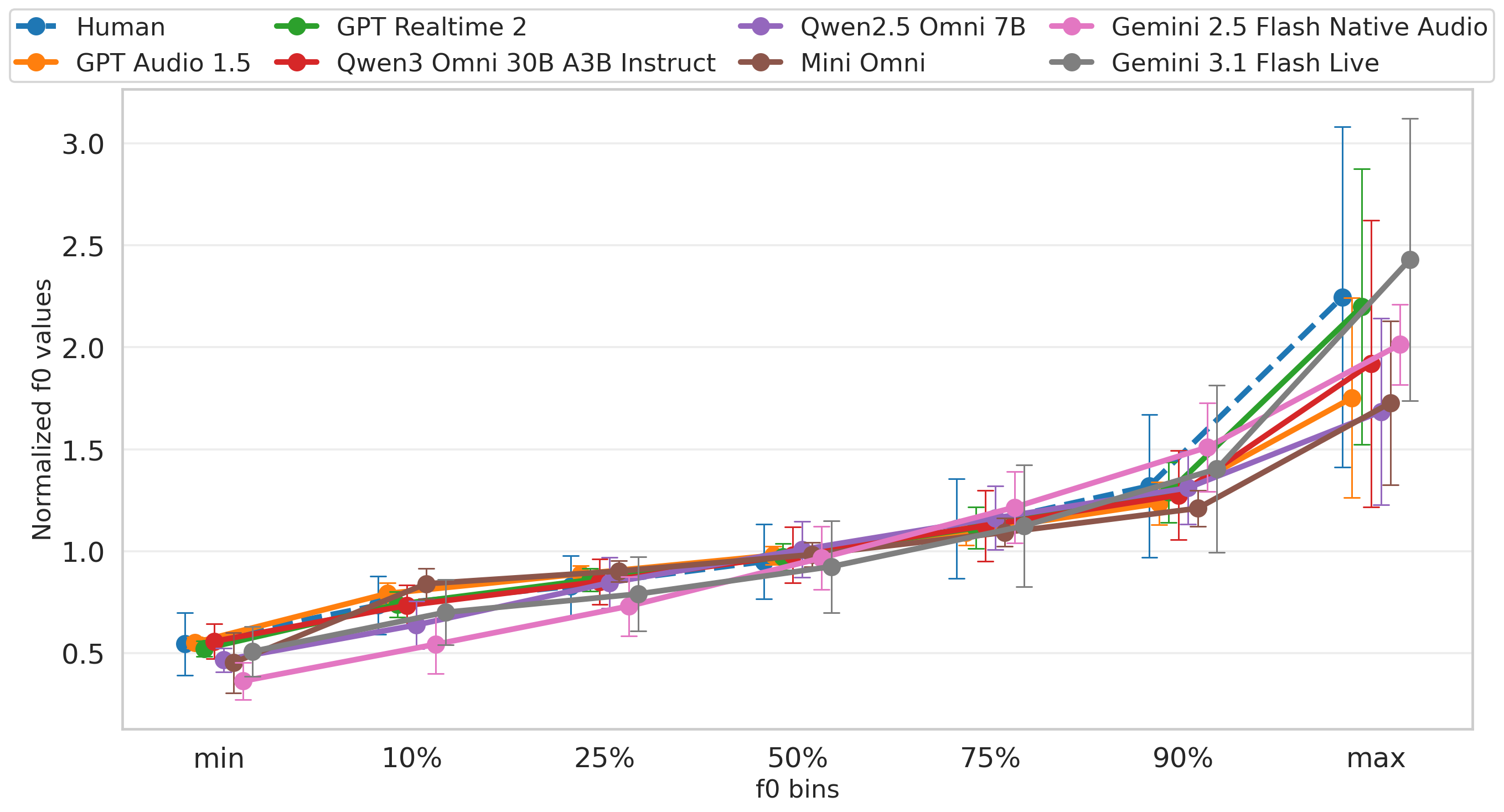}
    \caption{Distributions of the pitch features across all evaluated models and the human answers for multiple f0 bins.}
    \label{fig:S7}
\end{figure}

We use explainable temporal and spectral features to compare model and human speech in terms of pitch and timing. Figure~\ref{fig:S7} shows that model answers have a much smaller normalized pitch range than human answers across all bins and subsets, suggesting less expressive prosody. The last columns of Table \ref{tab:results} report this effect as \textit{Pitch Variation}, together with temporal features such as \textit{Answer Duration} and \textit{Voiced Ratio}. These results show that model answers are generally longer than human answers, while keeping a similar voice ratio.

\begin{table*}[bp]
\caption{\small SPEARBench results. Stance preservation, answer durations, and voiced ratio are reported as descriptive metrics. Best model results are reported in \textbf{bold}, second best in \textit{italic}. Interruptions are shown only for half- and full-duplex systems.}
\label{tab:results}
\centering
\scriptsize
\setlength{\tabcolsep}{3pt}
\resizebox{\linewidth}{!}{%
\begin{tabular}{l|ccc|ccc|c|ccc|cccc|ccc|ccc}
\toprule
 & \multicolumn{3}{c|}{\textbf{Speech quality}} & \multicolumn{3}{c|}{\textbf{Interruptions}} & \multicolumn{1}{c|}{\textbf{Turn-Taking}} & \multicolumn{3}{c|}{\textbf{Language and Dialect}} & \multicolumn{4}{c|}{\textbf{Emotions}} & \multicolumn{3}{c|}{\textbf{Stances}} & \multicolumn{3}{c}{\textbf{Explainable Features}} \\
\multirow{2}{*}{\textbf{Models}} & UTMOS & WER & CER & Latency & Interr. & Interr. & Naturalness & English & Dialectal & Dialectal & Emotional & Arousal & Valence & Dominance & Same & More & More & Answer & Voiced & Pitch \\
 &  &  &  &  & time & & average & answers & entrain. & variance & Naturalness & corr. & corr. & corr. & stance & negative & positive & Duration & ratio & Variation \\

 & 1-5$\uparrow$ & \%$\downarrow$ & \%$\downarrow$ & (ms)$\downarrow$ & (ms)$\downarrow$ & \%$\downarrow$ & score$\uparrow$ & \%$\uparrow$ & $\beta$$\uparrow$ & $tr(\Sigma)$$\uparrow$ & score$\uparrow$ & $\rho$$\uparrow$ & $\rho$$\uparrow$ & $\rho$$\uparrow$ & \% & \% & \% & (s) &  & (std)$\uparrow$ \\
\midrule
Human                                 & 2.22 & 27.1 & 17.2 & 742 & 521 & 33.8 & 4.99 & 99.3 & 0.53 & 250.6 & 11.10 & 0.53 & 0.36 & 0.51 & 97.1 & 1.5 & 1.7 & 8.8 & 0.500 & 0.321 \\
\midrule
\texttt{Gemini-2.5-flash-native-audio}& 3.72 & 7.9 & \textit{2.6} & 2137 & \textbf{0} & \textbf{0.0} & 5.31 & 99.7 & \textit{0.51} & 126.4 & 10.62 & \textit{0.19} & 0.27 & \textbf{0.15} & 97.1 & 1.3 & 1.8 & 6.6 & 0.514 & 0.161 \\
\texttt{Gemini-3.1-flash-live}       & 4.11 & 26.7 & 23.1         & 1154 & \textbf{0} & \textbf{0.0}  & \textbf{5.35} & 99.7 & \textbf{0.54} & \textit{178.0} & 10.16 & \textbf{0.25} & \textbf{0.29} & \textbf{0.20} & 97.6 & 1.7 & 0.6 & 10.6 & 0.568 & \textbf{0.299} \\
\texttt{GPT-audio-1.5}               & \textbf{4.36} & \textbf{5.8} & \textbf{1.4} & \textbf{959} & - & - & 4.74 & \textbf{100.0} & 0.46 & 104.2 & \textbf{10.91} & 0.11 & \textbf{0.29} & 0.09 & 98.0 & 0.8 & 1.2 & 9.8 & 0.548 & 0.118 \\
\texttt{GPT-realtime-2}              & 4.27 & 14.9 & 5.8 & 2182    & 431 & 23.6 & 5.33 & \textbf{100.0} & 0.41 & 90.0 & 10.89 & 0.06 & 0.21 & 0.08 & 97.5 & 1.5 & 0.9 & 19.3 & 0.508 & 0.164 \\
\texttt{Mini-Omni}                   & 3.92 & \textit{6.7} & 4.8  & 1243 & 1243 & 58.1 & 3.10 & \textbf{100.0} & 0.41 & 138.5 & 9.21 & -0.02 & 0.20 & -0.02 & 95.7 & 4.2 & 0.2 & 11.4 & 0.592 & 0.128 \\
\texttt{Qwen2.5-Omni-7B}             & 4.19 & 19.4 & 7.5          & \textit{1176} & \textbf{0} & \textbf{0.0} & 5.23 & 98.9 & 0.28 & 84.5 & 10.20 & -0.02 & -0.03 & -0.03 & 97.3 & 1.6 & 1.3 & 7.3 & 0.384 & 0.176 \\
\texttt{Qwen3-Omni-30B-A3B-Instruct} & \textit{4.34} & 7.9 & 3.3  & 2724 & \textbf{0} & \textbf{0.0} & \textit{5.34} & 98.1 & 0.48 & \textbf{204.5} & \textbf{10.91} & 0.07 & \textbf{0.29} & 0.06 & 97.2 & 1.1 & 2.0 & 8.6 & 0.671 & \textit{0.219} \\
\bottomrule
\end{tabular}
}
\end{table*}

\section{Discussion and Conclusion}

We introduced SPEARBench, an open benchmark and web platform for evaluating naturalness in speech-to-speech language models. SPEARBench builds two-turn question-answer interactions from the Seamless Interaction corpus, runs S2S model inference, and evaluates the generated answers with a unified set of automatic metrics covering intelligibility, speech quality, timing, interruptions, language and dialect consistency, emotional naturalness, interpersonal stance, turn-taking naturalness, and explainable distributional features. Each evaluation component is based on a previously validated model or metric, providing independent evidence of the reliability of the benchmark's measured dimensions.

The main goal of SPEARBench is to make conversational naturalness evaluation easy to run, compare, and extend. We release the curated evaluation data, the complete evaluation code, and the model-output reporting pipeline. The associated website provides a functional interface for browsing benchmark results, comparing systems across metrics, and supporting future model submissions. This makes the benchmark useful not only as a paper contribution, but also as a shared platform for tracking progress in speech-to-speech conversational modeling.

Our results show that current S2S LLMs can generate clean, intelligible speech but still differ from human conversational behavior in important ways. Models often show reduced pitch variation, longer answer durations, limited dialectal adaptation, weaker emotional entrainment, and, for some streaming systems, frequent interruptions. These findings confirm the need to evaluate S2S models not only as speech generators but also as conversational agents whose answers must be well-timed, expressive, emotionally appropriate, and socially consistent with the interaction.

The current benchmark has limitations: the evaluation is automatic, the data is English-only and two-speaker, timing measures depend partly on model serving interfaces, and pitch or dialect variability can be constrained by the voices available for each model. Future work will extend SPEARBench to more languages, conversational settings, and model submissions, while validating the platform with human judgments.

\newpage
\section{Generative AI Use Disclosure}
The authors used Claude (Anthropic) to assist with grammar checking and proofreading the manuscript. These tools were not used to generate any scientific content, results, or conclusions. All authors reviewed and take full responsibility for the final content of this paper.

\bibliography{main} 
\bibliographystyle{IEEEtran}

\end{document}